\documentclass[letterpaper, 10 pt, conference]{ieeeconf}  

\IEEEoverridecommandlockouts                              

\usepackage{amssymb,amsmath,amsfonts,mathrsfs}
\usepackage{graphicx}
\usepackage[table,xcdraw,dvipsnames]{xcolor}
\usepackage{tikz}
\usepackage{url}

\usepackage[utf8]{inputenc}

\usepackage[labelformat=simple]{subcaption}

\usepackage{colortbl}
\usepackage{booktabs}
\usepackage{array}
\usepackage{tabularx}

\usepackage[flushleft]{threeparttable}
\usepackage{multirow}

\let\labelindent\relax
\usepackage{enumitem}

\newcolumntype{L}[1]{>{\raggedright\let\newline\\\arraybackslash\hspace{0pt}}m{#1}}

\newcommand{\nth}[1]{\ensuremath{#1^\text{th}}}

\makeatletter
\DeclareRobustCommand\onedot{\futurelet\@let@token\@onedot}
\def\@onedot{\ifx\@let@token.\else.\null\fi\xspace}

\usepackage[T1]{fontenc}  
\usepackage[b]{esvect}    

\makeatletter
\newlength\xvec@height%
\newlength\xvec@depth%
\newlength\xvec@width%
\newcommand{\xvec}[2][]{%
  \ifmmode%
    \settoheight{\xvec@height}{$#2$}%
    \settodepth{\xvec@depth}{$#2$}%
    \settowidth{\xvec@width}{$#2$}%
  \else%
    \settoheight{\xvec@height}{#2}%
    \settodepth{\xvec@depth}{#2}%
    \settowidth{\xvec@width}{#2}%
  \fi%
  \def\xvec@arg{#1}%
  \def\xvec@dd{:}%
  \def\xvec@d{.}%
  \raisebox{.2ex}{\raisebox{\xvec@height}{\rlap{%
    \kern.05em
    \begin{tikzpicture}[scale=1]
    \pgfsetroundcap
    \draw (.05em,0)--(\xvec@width-.05em,0);
    \draw (\xvec@width-.05em,0)--(\xvec@width-.15em, .075em);
    \draw (\xvec@width-.05em,0)--(\xvec@width-.15em,-.075em);
    \ifx\xvec@arg\xvec@d%
      \fill(\xvec@width*.45,.5ex) circle (.5pt);%
    \else\ifx\xvec@arg\xvec@dd%
      \fill(\xvec@width*.30,.5ex) circle (.5pt);%
      \fill(\xvec@width*.65,.5ex) circle (.5pt);%
    \fi\fi%
    \end{tikzpicture}%
  }}}%
  #2%
}
\makeatother

\makeatletter
\renewcommand*\env@matrix[1][\arraystretch]{%
  \edef\arraystretch{#1}%
  \hskip -\arraycolsep
  \let\@ifnextchar\new@ifnextchar
  \array{*\c@MaxMatrixCols c}}
\makeatother

\definecolor{commentcolor}{gray}{0.5}
\usepackage{algorithm}
\usepackage{algpseudocode}
\algrenewcommand\algorithmicindent{1.0em}%

\algnewcommand{\LineComment}[1]{\State \textcolor{commentcolor}{\(\triangleright\) #1}}
\algnewcommand{\NewComment}[1]{\textcolor{commentcolor}{\(\triangleright\) #1}}
\algnewcommand{\To}{\textbf{to}}
\algnewcommand{\Break}{\textbf{break}}
\algnewcommand{\Continue}{\textbf{continue}}
\algnewcommand{\IIf}[1]{\State\algorithmicif\ #1\ \algorithmicthen}
\algnewcommand{\EndIIf}{\unskip}
\algnewcommand{\var}[1]{\textit{#1}}
\algnewcommand{\func}[1]{\textsc{#1}}

\title{\LARGE \bf
VTOL Failure Detection and Recovery by Utilizing Redundancy
}

\author{Mohammadreza Mousaei$^{1}$, 
Azarakhsh Keipour$^{2}$,
Junyi Geng$^{3}$
and Sebastian Scherer$^{4}$
\thanks{$^{1,3,4}$ Robotics Institute, Carnegie Mellon University, Pittsburgh, PA {\tt\small [mmousaei, junyigen, basti]@andrew.cmu.edu}}%
\thanks{$^{2}$ Robotics Institute, Carnegie Mellon University, Pittsburgh, PA {\tt\small keipour@cmu.edu, keipour@gmail.com}}%
}

\begin{document}

\maketitle
\thispagestyle{empty}
\pagestyle{empty}

\begin{abstract}

Offering vertical take-off and landing (VTOL) capabilities and the ability to travel great distances are crucial for Urban Air Mobility (UAM) vehicles. These capabilities make hybrid VTOLs the clear front-runners among UAM platforms.

On the other hand, concerns regarding the safety and reliability of autonomous aircraft have grown in response to the recent growth in aerial vehicle usage. As a result, monitoring the aircraft status to report any failures and recovering to prevent the loss of control when a failure happens are becoming increasingly important. Hybrid VTOLs can withstand some degree of actuator failure due to their intrinsic redundancy. Their aerodynamic performance, design, modeling, and control have all been addressed in the previous studies. However, research on their potential fault tolerance is still a less investigated field. 

In this workshop, we will present a summary of our work on aircraft fault detection and the recovery of our hybrid VTOL. First, we will go over our real-time aircraft-independent system for detecting actuator failures and abnormal behaviors. Then, in the context of our custom tiltrotor VTOL aircraft design, we talk about our optimization-based control allocation system, which utilizes the vehicle's configuration redundancy to recover from different actuation failures. Finally, we explore the ideas of how these parts can work together to provide a fail-safe system. We present our simulation and real-life experiments.

\end{abstract}


\section{Introduction} \label{sec:intro}

Fixed-rotor unmanned aerial vehicles (UAVs), such as multirotors, have vertical take-off and landing (VTOL) capabilities~\cite{keipour2020integration, keipour:mbz}. However, they are inefficient. On the other hand, fixed-wings are far more energy-efficient but lack VTOL capability~\cite{azarakhsh-icuas18}. Hybrid VTOL UAVs combine VTOL capabilities with efficient long-range flight and are the clear front-runners for Urban Air Mobility (UAM).

To securely incorporate UAVs into the airspace for real-world UAM applications, the aircraft should have a degree of tolerance to different hardware failures. While the controller is generally designed to be as robust to failures as possible~\cite{robust1, robust2}, failures may still happen. Hence, there is a need to detect and identify the failures and recover from them. 

\begin{figure}[!t]
    \centering
    \includegraphics[width=\linewidth]{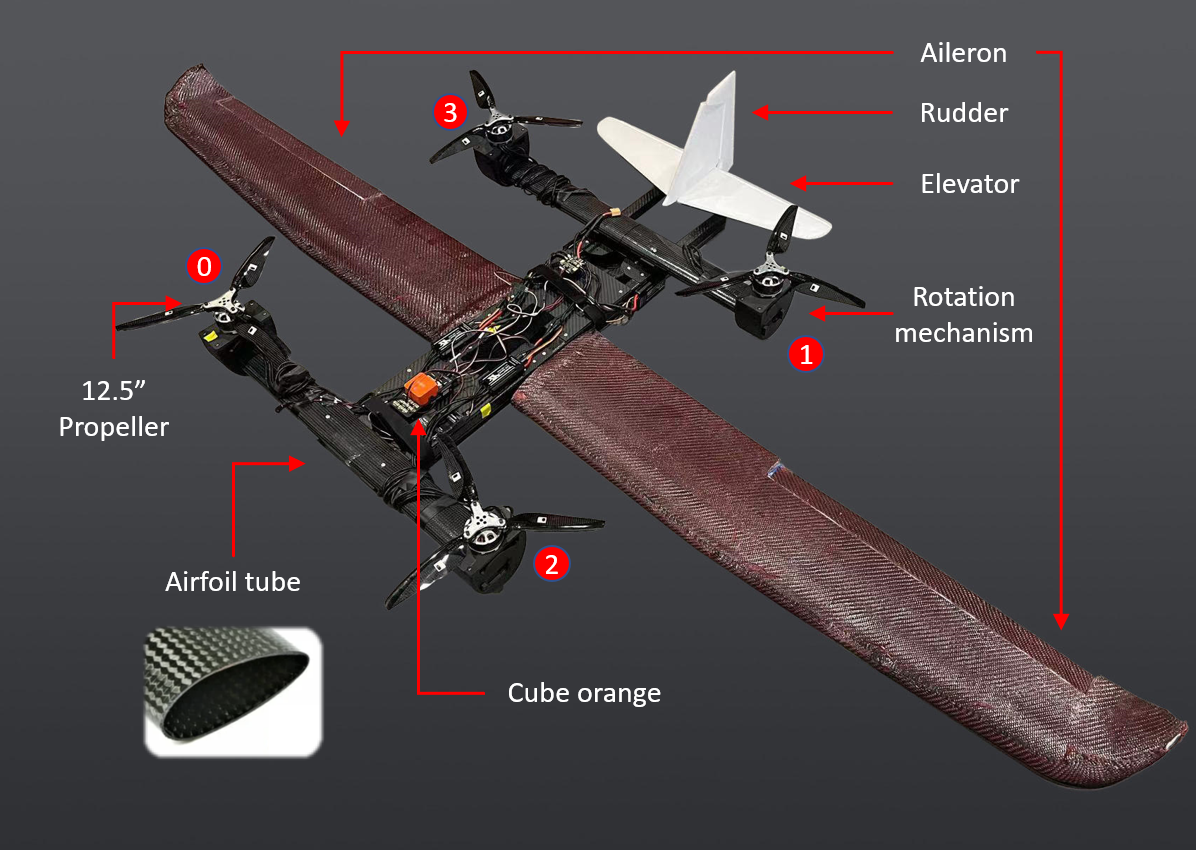}
    \caption{Our custom VTOL tiltrotor platform~\cite{mmousaeiIROS22}.}
    \label{fig:platform}
\end{figure}

The majority of the fault detection methods are heavily model-dependent~\cite{8014173, 6564801, AReviewonFault, 7526603}. On the other hand, signal processing-based techniques~\cite{UAVSecurity0167FrDTT2-04, birnbaum2015, 954410017691794} do not require the aircraft's model and instead detect the faults by analyzing the signals from the aircraft. 

On the failure recovery side, depending on the configuration of the aircraft, there are a variety of ways to deal with actuator failures~\cite{mueller2016relaxed, baskaya2021novel, stastny2018nonlinear}. In the case of hybrid VTOL UAVs, although there has been a recent increase in research~\cite{busan2021wind, lyu2017design, kamal2020conceptual, ducard2014modeling, zhang2018control, bauersfeld2021mpc}, fault tolerance has received only limited attention~\cite{fuhrer2019fault}. These aircraft offer configuration redundancy; however, little investigation has been done into using this redundancy to recover from actuator failures.

This workshop summarizes our works on aircraft fault detection~\cite{keipour2019automatic, azarakhsh-ijrr19} and failure recovery of our hybrid VTOL design~\cite{mmousaeiIROS22}. First, we describe our real-time RLS-based method for detecting actuator failures and anomalies in aircraft behavior. Instead of relying on the complete model of the aircraft, this approach models the connection between correlated input-output signal pairs and estimates a model online. The generated model is then used for real-time fault detection of virtually any aircraft without prior training. Then, we discuss tolerance to numerous types of actuator failures in the context of our custom tiltrotor VTOL aircraft design (Figure~\ref{fig:platform}). The aircraft is more resilient to actuator failures thanks to our designed dynamic control allocation, which takes advantage of system redundancy. The aircraft can thus recover from a collection of actuator failures in different flight phases by solving a constraint optimization problem. We present our simulation and real-world experiments performed to validate our methods. Finally, we explore the idea of how the detection and recovery systems can create a complete pipeline from detection to identification and recovery and discuss the possible future directions to achieve the safety requirement for UAM applications. 



\section{Our Work} \label{sec:method}


In this section, we first briefly describe our fault detection system, which can be used for almost any aircraft~\cite{keipour2019automatic}. Then we describe our custom VTOL system and discuss the fault recovery approach implemented in the context of this system~\cite{mmousaeiIROS22}. 

\begin{figure}[!t]
\centering
    \includegraphics[width=0.8\linewidth, height=0.3\textwidth]{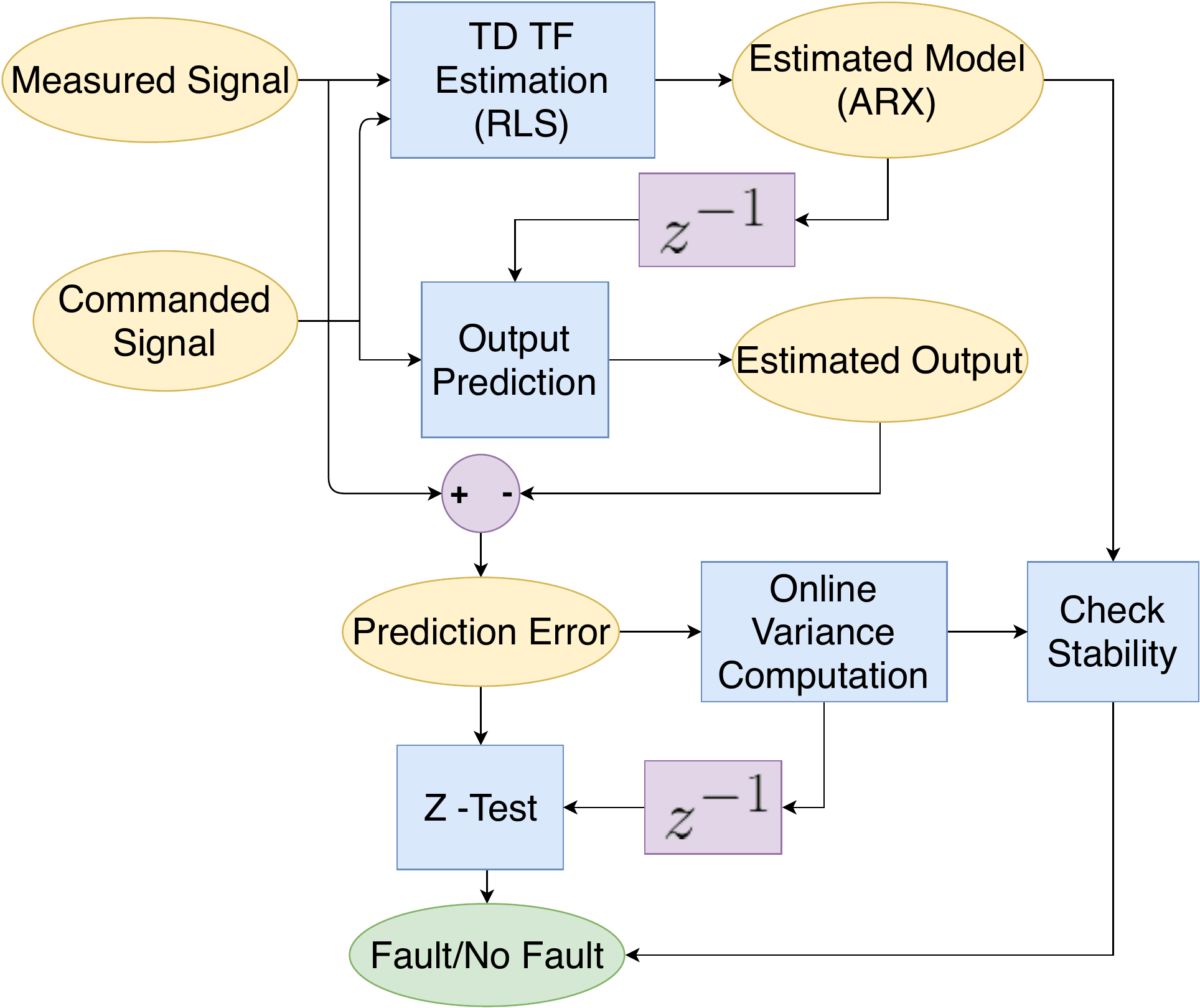}
    \caption{The flowchart of the fault detection method~\cite{keipour2019automatic}.}
    \label{fig:flowchart}
\end{figure}


The aircraft dynamics are nonlinear and cannot be described by a linear model.
However, instead of modeling the whole aircraft dynamics, the purpose of the detection part is to describe the interaction between two signals. Because many signal pairs are linearly connected to one another, we may use a linear model to estimate their connection. The aircraft dynamics were modeled using an Autoregressive Exogenous (ARX) Time-Domain Transfer Function model. Using our model, having past states and the current input is adequate to estimate the current output.

Assuming a true unknown ARX model for the relationship of input-output signals, the estimation algorithm aims to compute a prediction model which converges to the true model given enough samples. Recursive estimation methods aim to compute a new model estimate by a simple update to the current model when a new observation becomes available. 

Our approach~\cite{keipour2019automatic} uses the Recursive Least Squares (RLS) method, which is an online optimization method that recursively finds the coefficients to minimize a weighted linear least-squares cost function related to the input signals~\cite{hayes2009statistical}. Compared to most other methods, RLS exhibits fast convergence~\cite{1086206}. However, this benefit comes at the cost of higher computational complexity.

Once the ARX Time-Domain Transfer Function is estimated for the input-output pair, the output is predicted from the input using this model at each step. The error between the prediction and the measured output is utilized to update the model and calculate its Z-score. A high Z-score indicates a fault in the system. The algorithm can be written in the form of Fig.~\ref{fig:flowchart}.

We published our dataset consisting of tens of flights with various actuator failures on a real aircraft~\cite{azarakhsh-ijrr19}. It has been used in ours and several other works, including for failure identification with short identification times~\cite{identification}.


Based on the methods mentioned above, the failure can be detected and identified in near real-time. Given a correctly detected and identified failure, the next step is recovery from the failure. 

\begin{figure}[t]
    \centering
    \includegraphics[width=\linewidth]{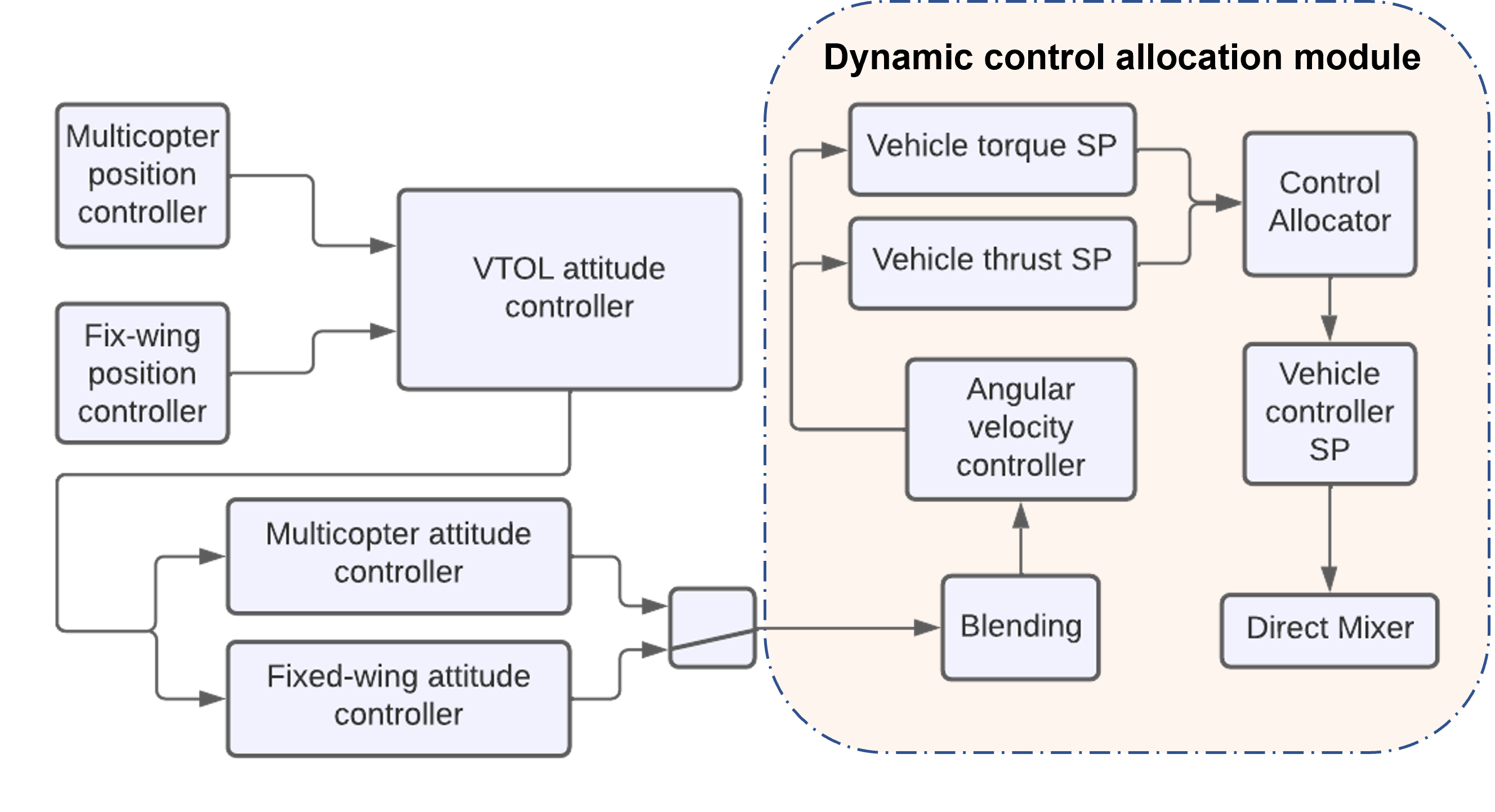}
    \caption{An overview of the VTOL recovery method~\cite{mmousaeiIROS22}.}
    \label{fig:control_arch}
\end{figure}

For our tiltrotor VTOL, we developed an optimization-based dynamic control allocation method in the PX4 flight control stack that can respond to the configuration changes of the vehicle, such as an actuator failure~\cite{mmousaeiIROS22}. Our designed control diagram is shown in Fig.~\ref{fig:control_arch}.

This control allocation module aims to allocate the desired control setpoint to the actuators depending on the current state and aircraft configuration. To execute this allocation, we define a constrained optimization problem. 

We first linearize the vehicle dynamics around the equilibrium points. This linearization leads to the control allocation matrix that maps our actuation change from the equilibrium points to the resulting forces and moments. To assign certain actuation for the desired set of forces and moments, the most straightforward technique is to use the pseudo-inverse of the control allocation matrix to obtain the least-norm solution.

An issue with the least-norm solution is that it might be out of actuation limits, which means that the desired values cannot be allocated to the actuators after trimming. This issue could cause the vehicle to deviate from the required forces and moments. Another problem is that consecutive solutions might be far apart, leading to jittery motor performance. So, even if the least-norm solution is feasible and within the range of actuation, it is not usable in practice.

To resolve the issues of the least-norm solution, we utilize the fact that our system is over-actuated, and there exists a non-zero null space that could be utilized to adjust the least-norm solution while still producing the desired set of forces and torques. We achieve this by combining the null space and the least-norm solutions. The least-norm solution achieves the net wrench, whereas the null space solution generates zero wrenches but remaps the inputs within the actuator limits while minimizing actuation change between consecutive steps. The final optimization problem is expressed as follows:
\begin{align} \label{eq:optimization}
    \min_{\boldsymbol{\lambda}} & J(\mathbf{u}_{\mathrm{sp}}) \\
    \mathrm{s.t} ~~&\mathbf{u}_{\mathrm{min},i}  \leq \mathbf{u}_{\mathrm{sp},i} \leq \mathbf{u}_{\mathrm{max},i} \nonumber
\end{align}
where $\mathbf{u}_{\mathrm{sp}}$ is the overall control setpoint, $J(\cdot)$ is the objective function where we try to minimize the actuator change from the trimmed condition, 
\begin{equation}
    J = (\mathbf{u}_{\mathrm{sp}} - \mathbf{u}_{\mathrm{sp}, \mathrm{trim}})^\top\mathbf{R}(\mathbf{u}_{\mathrm{sp}}-\mathbf{u}_{\mathrm{sp}, \mathrm{trim}})
\end{equation}
where $\mathbf{R}$ is the weight matrix considering the contribution from different actuators. The inequality constraints of problem (Eq.~\ref{eq:optimization}) ensure that the output is within actuator limits, where $i$ represents the $\nth{i}$ actuator. Simultaneously, minimizing the cost function ensures that the actuation change is minimized while the commands are kept near the equilibrium point. 

In the event of an actuation failure, we assume the failure has been detected and identified. With the knowledge of the actuation failure, we zero out or delete the corresponding column, which represents the failed actuator in the control allocation matrix. This provides the control allocation for the failure scenario. We subtract that from the desired wrench since the failed actuator still produces some wrench. The control allocation problem then becomes similar to that in the healthy vehicle. The only difference is that the problem dimension has been reduced by the number of failed actuators. Therefore, we formulate a similar constrained optimization problem for the vehicle with actuation failure.

\begin{table}
    \centering
    \caption{Actuator failure cases in different flight phases~\cite{mmousaeiIROS22}.}
    \label{tab:failcase} 
    \begin{threeparttable}
    \begin{tabular}{cll}
    \hline\hline
    Flight Phase                & Failure Description            & Cause                      \\
    \hline\hline
    \multirow{2}{*}{Multirotor} & Lock of one tilt in hover      & Broken servo               \\
                                & Single motor failure in hover  & Motor flaw/propeller loss  \\
    \hline
    \multirow{4}{*}{Fix-wing}   & Single motor failure in cruise & Motor flaw/propeller loss   \\
                                & Lock of one elevator in cruise & Broken servo               \\
                                & Lock of one aileron in cruise  & Broken servo  \\
    \hline\hline
    \end{tabular}
    \end{threeparttable}
\end{table}

\section{Experiments and Results} \label{sec:tests}

\subsection{Failure Detection}

\begin{table}
    \centering
    \caption{Failure detection statistics~\cite{keipour2019automatic}.}
    \label{tab:stats}
    \begin{threeparttable}
    \begin{tabular}{lccccc}
    \hline\hline
    \begin{tabular}[c]{@{}c@{}}Failure\\ Type\end{tabular}    & \begin{tabular}[c]{@{}c@{}}\# of \\ tests\end{tabular} & \begin{tabular}[c]{@{}c@{}}Flight\\ Time(s)\end{tabular} &  \begin{tabular}[c]{@{}c@{}}Avg.\\Detection\\Time(s)\end{tabular} & \begin{tabular}[c]{@{}c@{}}Max\\Detection\\Time(s)\end{tabular} &   \begin{tabular}[c]{@{}c@{}}Accuracy\\(\%)\end{tabular}\\
    \hline\hline
    Engine & 7 & 665 & 2.28 & 3.37 & 100 \\ \hline
    Rudder & 3 & 171 & 0.21 & 0.25 & -\\ \hline
    Elevator & 2 & 181 & 0.36 & 0.36 & -\\ \hline
    No Failure & 5 & 262 & - & - & -\\ \hline \hline
    \textbf{Total} & 22 & 1735 & 2.02 & 5.6 & 86.36\\ \hline\hline
    \end{tabular}
    \end{threeparttable}
\end{table}
Our failure detection system is developed and tested using C++ and ROS in Ubuntu on Nvidia Jetson TX2 onboard computer and Ardupilot autopilot is used on Pixhawk flight control. The flight test platform is a modified Carbon Z T-28, a fixed-wing UAV with a wingspan of 2 meters and a central electric engine.

The statistics from 22 flight tests with various types of failures are presented in Table~\ref{tab:stats}. Different performance metrics are used to evaluate our technique. Out of 22 tests, we found two False Positive (FP) and two False Negative (FN) detections, achieving $86.36$ percent accuracy, $88.23$ percent precision, and $88.23$ percent recall (sensitivity) within the total number of 19 correct sequences. A more detailed description of the tests and results is available in~\cite{keipour2019automatic} and our dataset article~\cite{azarakhsh-ijrr19} explains the assessment measures utilized for our calculations in further detail.

\begin{figure}[!t]
    \centering
    \includegraphics[width=0.6\linewidth]{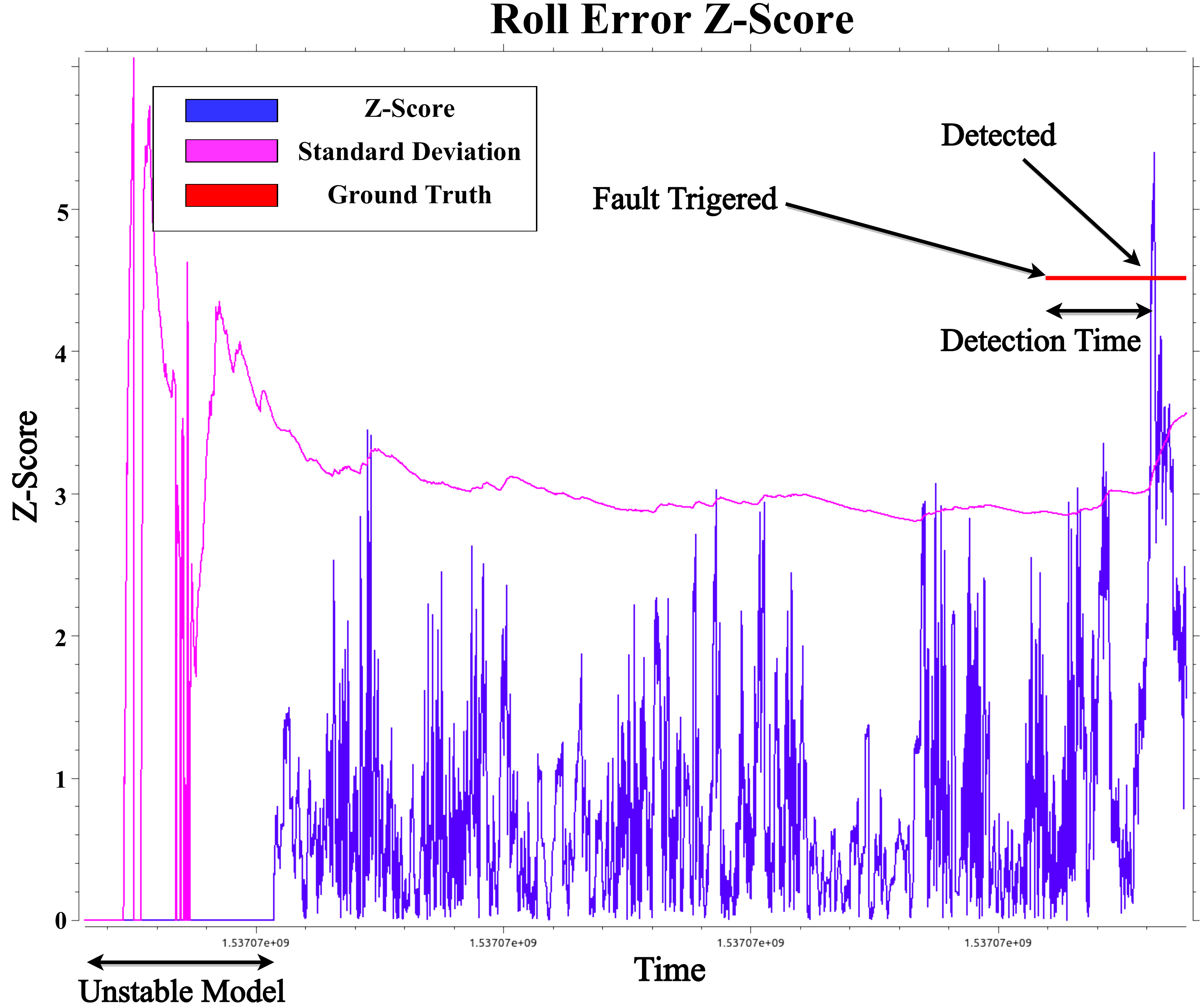}
    \caption{Z-Score vs Roll input for an engine failure flight test showing the detection of the fault from the sudden increase in Z-score~\cite{keipour2019automatic}.}
    \label{fig:roll-plot}
\end{figure}

Fig.~\ref{fig:roll-plot} shows how the system detects the engine failure from a pair of commanded/measured signals (roll error in this case). After the initial stabilization phase, the Z-score of the prediction error tends to be significantly less than the set threshold for the anomaly. However, when the failure happens, the prediction does not match the measurement anymore, and the spike in the Z-score indicates that an anomaly has happened. The figure also shows how the prediction error variance stabilizes with the stabilization of the predictor model. 

\subsection{Failure Recovery}
We model the tiltrotor VTOL in Gazebo based on our design in \cite{mmousaeiIROS22}. The dynamic control allocation is developed in PX4 autopilot, which can directly run on real aircraft. The constrained optimization is solved using Algilib~\cite{Algilib}, which is a numerical analysis and data processing library.

\subsubsection{Motor Failure in Hover}
\begin{figure}
    \centering
    \includegraphics[width=\linewidth]{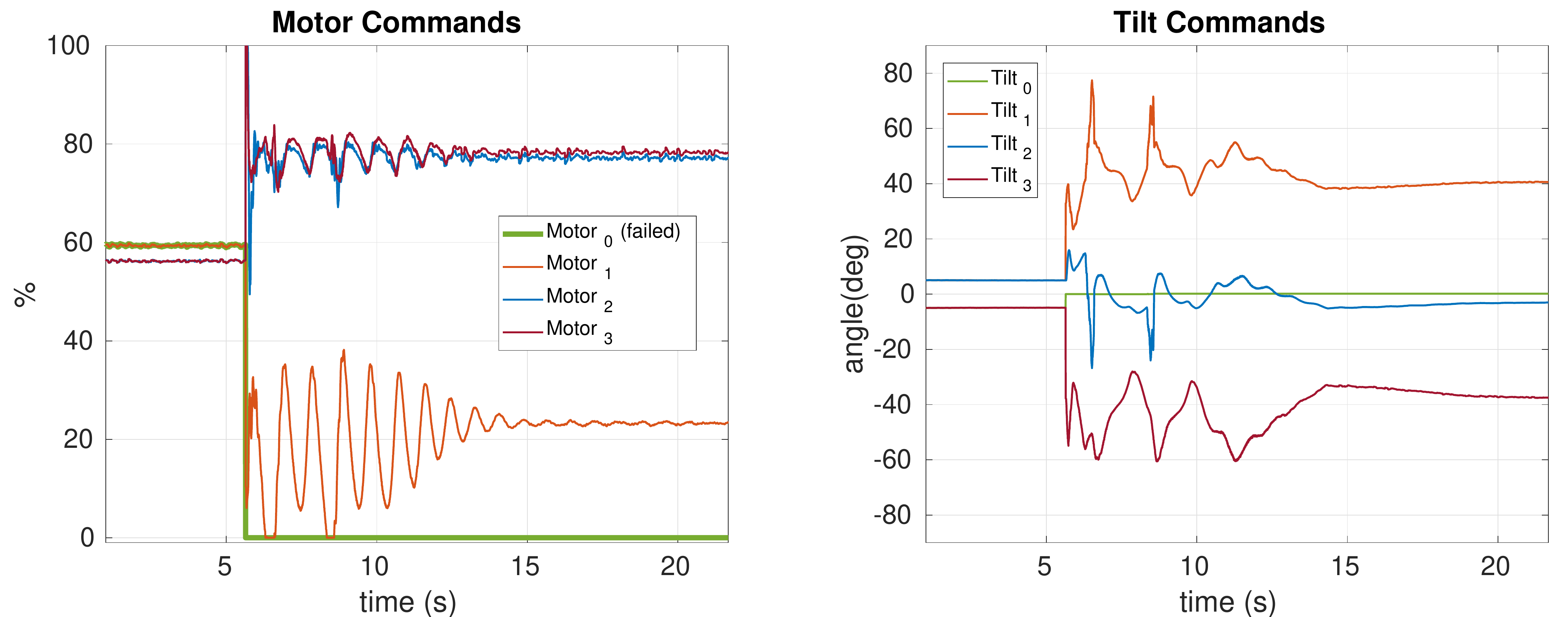}
    \caption{Actuator commands for a motor failure in hover~\cite{mmousaeiIROS22}.}
    \label{fig:motor_failure}
    \vspace{-2mm}
\end{figure}

When the vehicle takes off and hovers before transition, we completely turn down one of the motors to test this failure. The allocated actuator command is shown in Fig.~\ref{fig:motor_failure}. When the front-right motor fails, it is evident that all other effective actuators adjust and compensate for the thrust reduction. It is worth noting that the tilt angle associated with the failed motor immediately drops to zero, which is understandable given that the defective rotor can no longer create any applicable torques. After about 10 seconds, the actuator commands converge. After recovery, the aircraft is still controllable and can follow the planned waypoints.

\begin{figure}
    \centering
    \includegraphics[width=\linewidth]{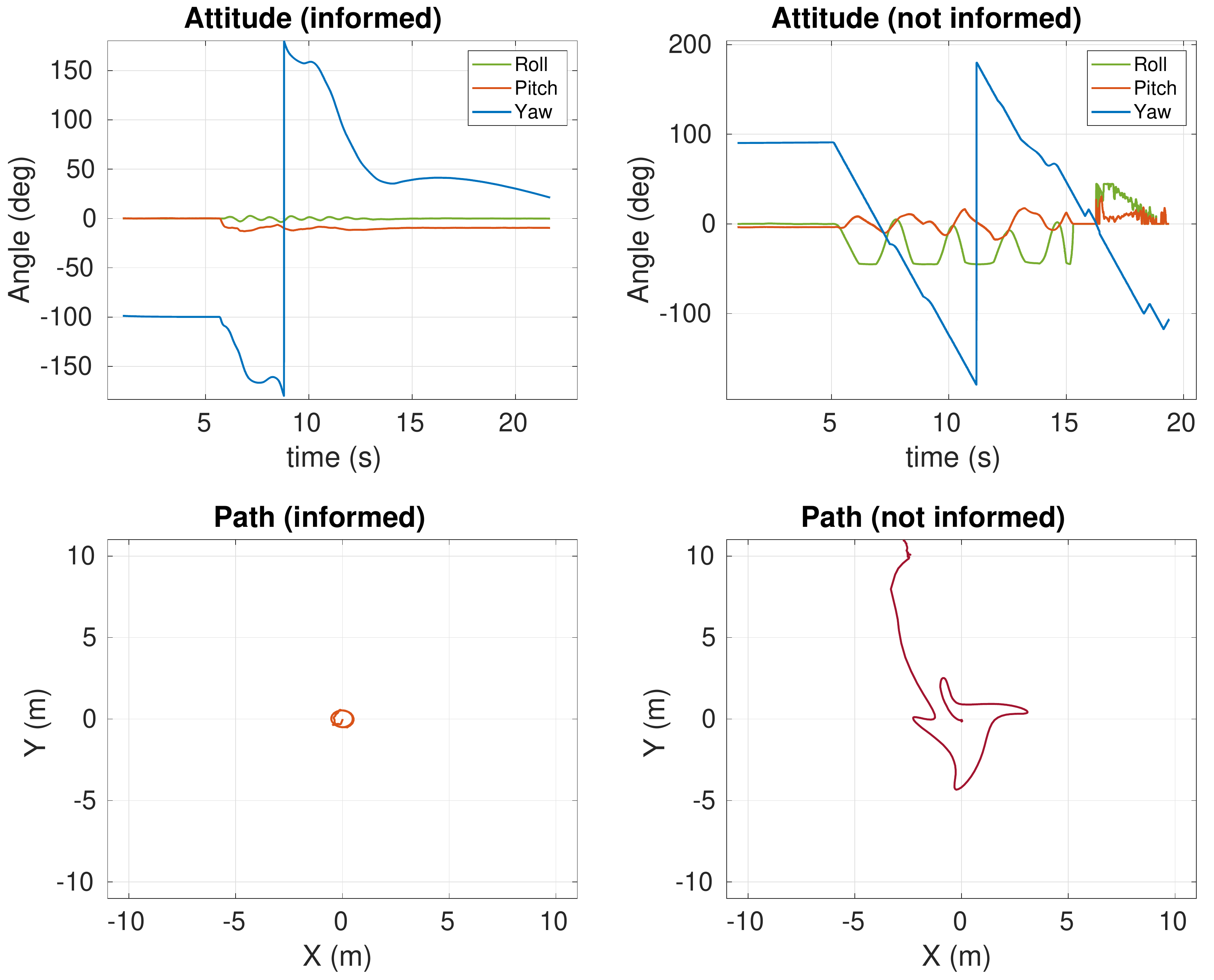}
    \caption{Aircraft attitude and path for a motor failure in hover~\cite{mmousaeiIROS22}.}
    \label{fig:attitude_path}
\end{figure}

We compare our approach to one in which the controller is unaware of the motor failure. The aircraft attitude and flight path are depicted in Fig.~\ref{fig:attitude_path} for two scenarios: with and without being alerted of the system failure. Without the knowledge of the failure, the system is still attempting to allocate a control wrench to the failed actuator. The vehicle instantly enters an extreme turn, eventually loses control, and crashes.

\subsubsection{Motor Failure During Cruise Flight}
\begin{figure}
    \centering
    \includegraphics[width=\linewidth]{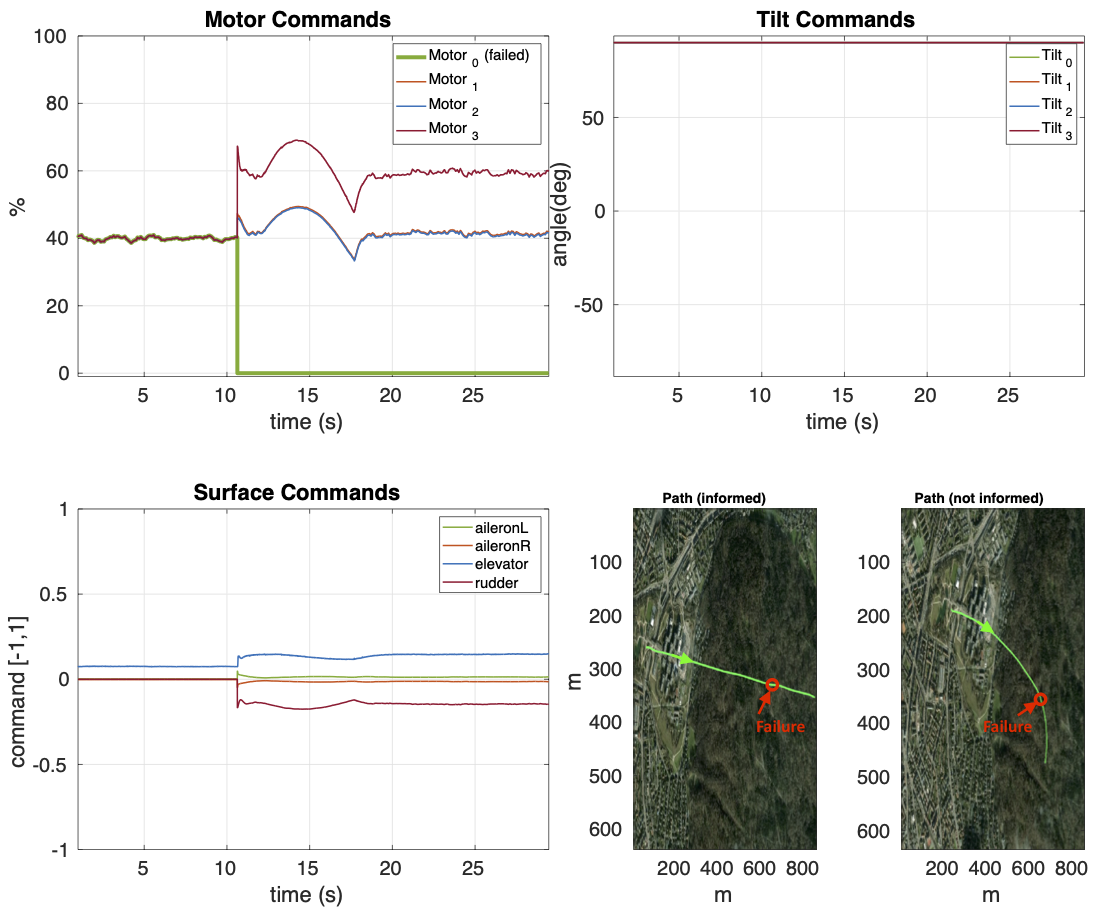}
    \caption{Actuator commands and flight path for a motor failure during cruise~\cite{mmousaeiIROS22}.}
    \label{fig:fw_motor_failure}
\end{figure}

Motor failure during the cruise flight is shown in Fig.~\ref{fig:fw_motor_failure}. The dynamical control allocation allows the aircraft to quickly adapt to the failure and resume normal flight. It is worth noting that the optimization solution maintains the complete tilts of all the rotors (all the rotors are facing forward). The vehicle can recover from the failure by altering the motor speed and control surfaces. In fact, in high-speed cruise flight, relying more on motor speed and control surfaces benefits the aircraft because an abrupt tilt change could be detrimental to the structure. Compared to the case in which the system is unaware of the failure, it is evident that the failure knowledge enables the aircraft to maintain a straight route, whereas being oblivious to the failure causes significant path deviation.

\subsubsection{Other Failures}
The list of failures tested with successful recovery is shown in Table. \ref{tab:failcase}. Tilt angle failure in the hover phase is simulated by injecting a sudden tilt servo lock at a fixed position (about 60$^{\circ}$ tilt). Even with a 60$^{\circ}$ tilt angle change, the system can recover quickly after a big sudden disturbance by dynamically reallocating the rest of the actuators. 

The elevator failure during cruise flight is injected with the control surface locked at 6$^{\circ}$. In this scenario, the rear two rotors tilt back, and the front two tilt slightly forward over 90$^{\circ}$ to compensate for the continual pitch up moment generated by the locked elevator while maintaining the force balanced. 

Finally, the aileron failure is injected during the cruise flight. When the aileron is locked at 15$^{\circ}$, the other healthy control surfaces adjust to compensate for the lower roll control authority in order to sustain cruise flight.

Ref.~\cite{mmousaeiIROS22} provides a more detailed description of the failure recovery experiments and the test results.

\section{Conclusions and Future Work} \label{sec:conclusion}

This workshop provided an overview of our failure detection and recovery progress. We described our method for real-time fault detection using a recursive least square algorithm and presented the real-world tests highlighting its accuracy of over 88 percent. Then, in the context of our custom tiltrotor VTOL, we explained our optimization-based control allocation approach for failure recovery. Finally, we showcased the extensive experimental results that validate our failure recovery capability under various actuator failures.

We suggest the following potential future work directions to move towards safer UAM vehicles:
\begin{itemize}
    \item Extending the detection system by monitoring multiple input vs. multiple output signals,
    \item Using a nonlinear model for failure detection to provide more precise predictions,
    \item Performing real flight tests to validate the described failure recovery approach further,
    \item Developing a failure identification systems to identify the type of the detected failure,
    \item Integrating the developed detection and recovery systems with an identification system to form a complete detection, identification, and recovery pipeline.
\end{itemize}

\addtolength{\textheight}{-5.6cm}   



\section*{Acknowledgment}

The authors would like to thank Dongwei Bai for his support and help with the mechanical build of the vehicle.


\bibliographystyle{IEEEtran}
\bibliography{paper-citations.bib}

\begin{thebibliography}{10}
\providecommand{\url}[1]{#1}
\csname url@rmstyle\endcsname
\providecommand{\newblock}{\relax}
\providecommand{\bibinfo}[2]{#2}
\providecommand\BIBentrySTDinterwordspacing{\spaceskip=0pt\relax}
\providecommand\BIBentryALTinterwordstretchfactor{4}
\providecommand\BIBentryALTinterwordspacing{\spaceskip=\fontdimen2\font plus
\BIBentryALTinterwordstretchfactor\fontdimen3\font minus
  \fontdimen4\font\relax}
\providecommand\BIBforeignlanguage[2]{{%
\expandafter\ifx\csname l@#1\endcsname\relax
\typeout{** WARNING: IEEEtran.bst: No hyphenation pattern has been}%
\typeout{** loaded for the language `#1'. Using the pattern for}%
\typeout{** the default language instead.}%
\else
\language=\csname l@#1\endcsname
\fi
#2}}

\bibitem{keipour2020integration}
A.~Keipour, M.~Mousaei, A.~T. Ashley, and S.~Scherer, ``Integration of
  fully-actuated multirotors into real-world applications,'' \emph{arXiv
  preprint arXiv:2011.06666}, 2020.

\bibitem{keipour:mbz}
\BIBentryALTinterwordspacing
A.~Keipour, G.~A.~S. Pereira, R.~Bonatti, R.~Garg, P.~Rastogi, G.~Dubey, and
  S.~Scherer, ``Visual servoing approach for autonomous {UAV} landing on a
  moving vehicle,'' \emph{arXiv:2104.01272}, 2021. [Online]. Available:
  \url{https://arxiv.org/abs/2104.01272}
\BIBentrySTDinterwordspacing

\bibitem{azarakhsh-icuas18}
S.~Schopferer, J.~S. Lorenz, A.~Keipour, and S.~Scherer, ``Path planning for
  unmanned fixed-wing aircraft in uncertain wind conditions using trochoids,''
  in \emph{2018 International Conference on Unmanned Aircraft Systems (ICUAS)},
  June 2018, pp. 503--512.

\bibitem{robust1}
N.~Hegde, V.~George, C.~{Gurudas Nayak}, and K.~Kumar,
  ``\BIBforeignlanguage{English}{Transition flight modeling and robust control
  of a {VTOL} unmanned quad tilt-rotor aerial vehicle},''
  \emph{\BIBforeignlanguage{English}{Indonesian Journal of Electrical
  Engineering and Computer Science}}, vol.~18, no.~3, pp. 1252--1261, Jan.
  2020.

\bibitem{robust2}
J.~A. Guerrero, R.~Lozano, G.~Romero, D.~Lara-Alabazares, and K.~C. Wong,
  ``Robust control design based on sliding mode control for hover flight of a
  mini tail-sitter unmanned aerial vehicle,'' in \emph{2009 35th Annual
  Conference of IEEE Industrial Electronics}, 2009, pp. 2342--2347.

\bibitem{mmousaeiIROS22}
\BIBentryALTinterwordspacing
M.~Mousaei, J.~Geng, A.~Keipour, D.~Bai, and S.~Scherer, ``Design, modeling and
  control for a tilt-rotor vtol uav in the presence of actuator failure,''
  2022. [Online]. Available: \url{https://arxiv.org/abs/2205.05533}
\BIBentrySTDinterwordspacing

\bibitem{8014173}
M.~Kuric, B.~Lacevic, N.~Osmic, and A.~Tahirovic, ``Rls-based fault-tolerant
  tracking control of multirotor aerial vehicles,'' in \emph{2017 IEEE
  International Conference on Advanced Intelligent Mechatronics (AIM)}, July
  2017, pp. 1148--1153.

\bibitem{6564801}
X.~Qi, D.~Theilliol, J.~Qi, Y.~Zhang, and J.~Han, ``A literature review on
  fault diagnosis methods for manned and unmanned helicopters,'' in \emph{2013
  International Conference on Unmanned Aircraft Systems (ICUAS)}, May 2013, pp.
  1114--1118.

\bibitem{AReviewonFault}
\BIBentryALTinterwordspacing
X.~Qi, J.~Qi, D.~Theilliol, Y.~Zhang, J.~Han, D.~Song, and C.~Hua, ``A review
  on fault diagnosis and fault tolerant control methods for single-rotor aerial
  vehicles,'' \emph{Journal of Intelligent {\&} Robotic Systems}, vol.~73,
  no.~1, pp. 535--555, Jan 2014. [Online]. Available:
  \url{https://doi.org/10.1007/s10846-013-9954-z}
\BIBentrySTDinterwordspacing

\bibitem{7526603}
A.~Ansari and D.~S. Bernstein, ``Aircraft sensor fault detection using state
  and input estimation,'' in \emph{2016 American Control Conference (ACC)},
  July 2016, pp. 5951--5956.

\bibitem{UAVSecurity0167FrDTT2-04}
Z.~Birnbaum, A.~Dolgikh, V.~Skormin, E.~O'Brien, D.~Muller, and
  C.~Stracquodaine, ``Unmanned aerial vehicle security using behavioral
  profiling,'' in \emph{2015 International Conference on Unmanned Aircraft
  Systems (ICUAS)}, June 2015, pp. 1310--1319.

\bibitem{birnbaum2015}
\BIBentryALTinterwordspacing
Z.~Birnbaum, A.~Dolgikh, V.~Skormin, E.~O'Brien, and D.~Muller, ``Unmanned
  aerial vehicle security using recursive parameter estimation,'' \emph{Journal
  of Intelligent {\&} Robotic Systems}, vol.~84, no.~1, pp. 107--120, Dec 2016.
  [Online]. Available: \url{https://doi.org/10.1007/s10846-015-0284-1}
\BIBentrySTDinterwordspacing

\bibitem{954410017691794}
\BIBentryALTinterwordspacing
W.~Han, Z.~Wang, and Y.~Shen, ``Fault estimation for a quadrotor unmanned
  aerial vehicle by integrating the parity space approach with recursive least
  squares,'' \emph{Proceedings of the Institution of Mechanical Engineers, Part
  G: Journal of Aerospace Engineering}, vol. 232, no.~4, pp. 783--796, 2018.
  [Online]. Available: \url{https://doi.org/10.1177/0954410017691794}
\BIBentrySTDinterwordspacing

\bibitem{mueller2016relaxed}
M.~W. Mueller and R.~D’Andrea, ``Relaxed hover solutions for multicopters:
  Application to algorithmic redundancy and novel vehicles,'' \emph{The
  International Journal of Robotics Research}, vol.~35, no.~8, pp. 873--889,
  2016.

\bibitem{baskaya2021novel}
E.~Baskaya, M.~Hamandi, M.~Bronz, and A.~Franchi, ``A novel robust hexarotor
  capable of static hovering in presence of propeller failure,'' \emph{IEEE
  Robotics and Automation Letters}, vol.~6, no.~2, pp. 4001--4008, 2021.

\bibitem{stastny2018nonlinear}
T.~Stastny and R.~Siegwart, ``Nonlinear model predictive guidance for
  fixed-wing {UAV}s using identified control augmented dynamics,'' in
  \emph{2018 International Conference on Unmanned Aircraft Systems
  (ICUAS)}.\hskip 1em plus 0.5em minus 0.4em\relax IEEE, 2018, pp. 432--442.

\bibitem{busan2021wind}
R.~C. Busan, P.~C. Murphy, D.~B. Hatke, and B.~M. Simmons, ``Wind tunnel
  testing techniques for a tandem tilt-wing, distributed electric propulsion
  {VTOL} aircraft,'' in \emph{AIAA SciTech 2021 Forum}, 2021, p. 1189.

\bibitem{lyu2017design}
X.~Lyu, H.~Gu, Y.~Wang, Z.~Li, S.~Shen, and F.~Zhang, ``Design and
  implementation of a quadrotor tail-sitter {VTOL} {UAV},'' in \emph{2017 IEEE
  international conference on robotics and automation (ICRA)}.\hskip 1em plus
  0.5em minus 0.4em\relax IEEE, 2017, pp. 3924--3930.

\bibitem{kamal2020conceptual}
A.~Kamal and A.~Ramirez-Serrano, ``Conceptual design of a highly-maneuverable
  transitional {VTOL} {UAV} with new maneuver and control capabilities,'' in
  \emph{AIAA Scitech 2020 Forum}, 2020, p. 1733.

\bibitem{ducard2014modeling}
G.~Ducard and M.-D. Hua, ``Modeling of an unmanned hybrid aerial vehicle,'' in
  \emph{2014 IEEE Conference on Control Applications (CCA)}.\hskip 1em plus
  0.5em minus 0.4em\relax IEEE, 2014, pp. 1011--1016.

\bibitem{zhang2018control}
J.~Zhang, P.~Bhardwaj, S.~A. Raab, S.~Saboo, and F.~Holzapfel, ``Control
  allocation framework for a tilt-rotor vertical take-off and landing
  transition aircraft configuration,'' in \emph{2018 Applied Aerodynamics
  Conference}, 2018, p. 3480.

\bibitem{bauersfeld2021mpc}
L.~Bauersfeld, L.~Spannagl, G.~J. Ducard, and C.~H. Onder, ``Mpc flight control
  for a tilt-rotor {VTOL} aircraft,'' \emph{IEEE Transactions on Aerospace and
  Electronic Systems}, vol.~57, no.~4, pp. 2395--2409, 2021.

\bibitem{fuhrer2019fault}
S.~Fuhrer, S.~Verling, T.~Stastny, and R.~Siegwart, ``Fault-tolerant flight
  control of a {VTOL} tailsitter {UAV},'' in \emph{2019 International
  Conference on Robotics and Automation (ICRA)}.\hskip 1em plus 0.5em minus
  0.4em\relax IEEE, 2019, pp. 4134--4140.

\bibitem{keipour2019automatic}
\BIBentryALTinterwordspacing
A.~Keipour, M.~Mousaei, and S.~Scherer, ``Automatic real-time anomaly detection
  for autonomous aerial vehicles,'' in \emph{2019 International Conference on
  Robotics and Automation (ICRA)}.\hskip 1em plus 0.5em minus 0.4em\relax IEEE,
  2019, pp. 5679--5685. [Online]. Available:
  \url{https://ieeexplore.ieee.org/document/8794286}
\BIBentrySTDinterwordspacing

\bibitem{azarakhsh-ijrr19}
\BIBentryALTinterwordspacing
------, ``{ALFA}: A dataset for {UAV} fault and anomaly detection,'' \emph{The
  International Journal of Robotics Research}, vol.~40, no. 2-3, pp. 515--520,
  2021. [Online]. Available:
  \url{https://journals.sagepub.com/doi/10.1177/0278364920966642}
\BIBentrySTDinterwordspacing

\bibitem{hayes2009statistical}
M.~H. Hayes, \emph{Statistical digital signal processing and modeling}.\hskip
  1em plus 0.5em minus 0.4em\relax John Wiley \& Sons, 2009.

\bibitem{1086206}
E.~Eweda and O.~Macchi, ``Convergence of the rls and lms adaptive filters,''
  \emph{IEEE Transactions on Circuits and Systems}, vol.~34, no.~7, pp.
  799--803, July 1987.

\bibitem{identification}
\BIBentryALTinterwordspacing
M.~W. Ahmad, M.~U. Akram, R.~Ahmad, K.~Hameed, and A.~Hassan, ``Intelligent
  framework for automated failure prediction, detection, and classification of
  mission critical autonomous flights,'' \emph{ISA Transactions}, 2022.
  [Online]. Available:
  \url{https://www.sciencedirect.com/science/article/pii/S0019057822000209}
\BIBentrySTDinterwordspacing

\bibitem{Algilib}
\BIBentryALTinterwordspacing
S.~Bochkano. (1999) Alglib. Accessed: 2022-03-01. [Online]. Available:
  \url{www.alglib.net}
\BIBentrySTDinterwordspacing

\end{thebibliography}

\end{document}